\documentclass[letterpaper]{article}
\usepackage[preprint]{aaai2027}
\usepackage[hyphens]{url}
\usepackage{graphicx}
\usepackage{natbib}
\usepackage{caption}
\usepackage{amsmath}
\usepackage{amssymb}
\usepackage{algorithm}
\usepackage{algorithmic}
\usepackage{booktabs}

\title{LIRA: A More Interpretable Dynamic Loss-Informed Rank Allocation for Low-Rank Adaptation}
\author{Sihao Wang\equalcontrib \quad Bozheng Li\equalcontrib \quad Xuemiao Xu \\
        Haoxin Yang \quad Yujie Xu \quad Jie Tang}
\affiliations{South China University of Technology}

\begin{document}
\maketitle

\begin{abstract}
Low-rank adaptation (LoRA) has become a widely used parameter-efficient fine-tuning method for large pretrained models.
However, standard LoRA typically assigns a fixed rank profile despite layer- and stage-dependent adaptation needs, while recent adaptive-rank variants often rely on heuristic or first-order signals that provide only an indirect link between rank reallocation and local loss behavior. This leaves the rationale behind each rank reallocation insufficiently interpretable.
Motivated by this gap, we propose \textbf{LIRA} (\textbf{L}oss-\textbf{I}nformed \textbf{R}ank \textbf{A}llocation), a dynamic-rank LoRA method designed to make online rank reallocation \emph{more interpretable}.
LIRA scores each layer by the estimated magnitude of the local loss perturbation associated with its recent LoRA-factor movements under a second-order approximation, making the basis for each rank change interpretable term by term.
By periodically applying \textbf{DELTA} (\textbf{D}isplacement-\textbf{E}valuated \textbf{L}oss-\textbf{T}rajectory \textbf{A}llocation), a displacement-based algorithm for LoRA rank reallocation, LIRA produces discrete effective ranks under matched budgets while remaining compatible with standard $(A,B)$ adapters.
On RoBERTa-base GLUE under the La-LoRA/LoRA protocol, LIRA outperforms the uniform-LoRA and adaptive-rank baselines reported under this protocol; Trajectory EMA attains performance competitive with Fixed-point Fisher at lower estimation cost.
\end{abstract}

\section{Introduction}
\label{sec:intro}

Pretrained language models have become the dominant starting point for downstream natural language processing~\citep{devlin2019bert,brown2020language}.
The standard way to adapt such models is to fine-tune all parameters on task-specific data.
However, as model scale grows, full fine-tuning incurs substantial memory and storage costs, especially when many downstream tasks must be served concurrently and each task would otherwise require a separate full model copy.

To address this issue, researchers have proposed parameter-efficient fine-tuning (PEFT) methods that freeze the base model and update only a small set of task-specific parameters~\citep{houlsby2019parameter,li2021prefix}.
Low-rank adaptation (LoRA)~\citep{hu2022lora} is a representative PEFT method: standard LoRA typically prespecifies a fixed rank profile for all adapted layers.
This approach implicitly assumes that different layers require similar adaptation capacity throughout fine-tuning.

\begin{table}[t]
\centering
\small
\setlength{\tabcolsep}{1mm}
\begin{tabular*}{\columnwidth}{@{\extracolsep{\fill}}p{0.19\columnwidth}p{0.25\columnwidth}p{0.48\columnwidth}@{}}
\toprule
{\raggedright Method\par} & {\raggedright Allocation unit\par} & {\raggedright Importance signal\par} \\
\midrule
AdaLoRA &
{\raggedright Singular-value triplets\par} &
{\raggedright Parameter--gradient sensitivity\par} \\
\addlinespace[1pt]

IGU-LoRA &
{\raggedright Singular-value triplets\par} &
{\raggedright Integrated gradients with uncertainty-aware smoothing\par} \\
\addlinespace[1pt]

La-LoRA &
{\raggedright Layer-wise rank\par} &
{\raggedright Norm growth / layer contribution\par} \\
\addlinespace[1pt]

SoRA &
{\raggedright Rank dimension\par} &
{\raggedright Sparse gates\par} \\
\addlinespace[1pt]

L1RA &
{\raggedright Rank dimension\par} &
{\raggedright $L_1$-regularized sparse gates\par} \\
\addlinespace[1pt]

ALoRA &
{\raggedright Rank dimension\par} &
{\raggedright Validation-time per-rank ablation\par} \\
\addlinespace[1pt]

\textbf{LIRA} &
{\raggedright\textbf{Layer-wise effective rank}\par} &
{\raggedright\textbf{Local loss-perturbation magnitude of $\Delta A,\Delta B$, normalized by $r_i$}\par} \\
\bottomrule
\end{tabular*}
\caption{Online rank-allocation methods, allocation units, and importance signals.}
\label{tab:related-comparison}
\end{table}

\begin{figure}[t]
\centering
\includegraphics[width=\linewidth]{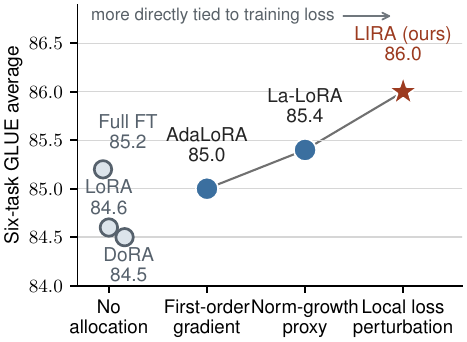}
\caption{Six-task GLUE averages by rank-allocation signal. Groups are descriptive rather than an interpretability scale. Baselines are from La-LoRA~\citep{gu2026lalora}.}
\label{fig:interp-vs-acc}
\end{figure}

In practice, however, adaptation needs can vary across modules, tasks, and training stages, so a fixed profile may under-supply some layers while spending capacity on others.
This potential mismatch has motivated methods that allocate ranks or budgets dynamically during training.
For example, AdaLoRA~\citep{zhang2023adalora} adjusts singular-value triplets according to parameter--gradient sensitivity, whereas La-LoRA~\citep{gu2026lalora} allocates layer-wise ranks based on norm growth and layer contribution.
Table~\ref{tab:related-comparison} summarizes representative methods with their allocation units and importance signals.

Despite these advances, Figure~\ref{fig:interp-vs-acc} shows that existing methods mainly allocate ranks using proxy signals such as first-order gradients, parameter sensitivity, and norm growth.
Although these signals can reflect parameter importance, they do not directly explain how assigning more rank to a highly scored layer improves the training loss.
In other words, proxy scores, the resulting rank allocation, and loss improvement still lack an intuitive mathematical connection, which may limit the effective use of a finite rank budget and make the allocation decision difficult to interpret in terms of the training objective.

Motivated by this gap, we propose \textbf{LIRA} (\textbf{L}oss-\textbf{I}nformed \textbf{R}ank \textbf{A}llocation), a dynamic-rank LoRA method for online rank reallocation.
Under a local second-order approximation, LIRA brings observed LoRA-factor displacements, gradients, and curvature together in an estimate of local loss perturbation, thereby establishing an intuitive mathematical connection between rank allocation and loss optimization.
After rank normalization, the resulting perturbation magnitude remains expressed in loss units and can be interpreted as the local loss effect per effective rank of a layer; LIRA then uses this score to reallocate effective ranks.
Consequently, each rank adjustment can be traced to specific terms in the loss approximation, making the basis for allocation more interpretable.

\noindent Our main contributions are as follows:
\begin{itemize}
  \item For \textbf{rank scoring interpretability}, we derive rank-normalized influence scores from a local second-order loss approximation induced by recent LoRA-factor movements; these scores can be estimated using lightweight training-trajectory statistics or an optional fixed-point Fisher estimate.
  \item For \textbf{online LoRA rank allocation}, we propose \textbf{DELTA} (\textbf{D}isplacement-\textbf{E}valuated \textbf{L}oss-\textbf{T}rajectory \textbf{A}llocation), which adapts Fisher-inspired influence estimation to the fine-tuning process by using either successive mini-batches or optimization-step trajectories as natural samples, and uses the resulting scores to update stable integer per-layer ranks through temperature-softmax targets and discrete projection.
  \item For \textbf{empirical validation}, under the La-LoRA/LoRA RoBERTa-base GLUE protocol, the two LIRA instantiations---Trajectory EMA and Fixed-point Fisher---achieve six-task averages of $86.0$ and $86.1$, respectively, both surpassing the cited uniform-LoRA baseline of $84.6$ and La-LoRA baseline of $85.4$; Trajectory EMA requires no additional statistic-window computation.
\end{itemize}

\section{Related Work}
\label{sec:related}

\subsection{Low-Rank Adaptation with Fixed Rank Profiles}
Low-rank adaptation (LoRA)~\citep{hu2022lora} keeps pretrained weights frozen and learns a low-rank increment $\Delta W=BA$ for downstream adaptation.
Subsequent work refines how this increment is formed or optimized: LoRA+ uses asymmetric learning rates for the two factors~\citep{hayou2024lora_plus}, DoRA applies magnitude--direction decomposition~\citep{liu2024dora}, and VeRA combines shared random bases with a small set of trainable coefficients~\citep{kopiczko2024vera}.
Although these methods improve expressiveness or training efficiency, they share one premise: the per-layer rank profile remains fixed throughout fine-tuning, with no active rebalancing of capacity across layers.

Unlike these methods, we do not fix the per-layer rank profile before fine-tuning. Instead, under a limited total rank budget, we dynamically reallocate effective ranks across layers according to the estimated magnitude of each layer's local loss perturbation.
We retain standard $(A,B)$ LoRA and control only each layer's effective rank---the number of rank components that actually participate in the forward pass.
This design adds little extra overhead, fits existing fine-tuning pipelines, and enables fair comparison with uniform-rank LoRA under a matched total budget.

\subsection{Online Dynamic Rank and Budget Allocation}
Uniformly spreading a limited rank budget across all layers is generally suboptimal.
Prior observations show that, under a fixed parameter budget, adapting FFN weights or top layers can outperform adapting attention modules or bottom layers~\citep{zhang2023adalora}.
Existing work also explores non-uniform rank profiles at different stages and for different objectives.
FIM-LoRA~\citep{fimlora2026}, GoRA~\citep{he2025gora}, and GeLoRA~\citep{eddib2025gelora} profile importance before regular fine-tuning, after which their rank configurations are largely static; DyLoRA~\citep{valipour2023dylora} instead targets multi-rank compatibility within a single adapter.

Unlike the above approaches, which determine rank configurations before regular fine-tuning, we focus on online dynamic rank allocation, i.e., adjusting rank or effective capacity during training according to layer importance.
Existing online dynamic-rank methods differ primarily in how they construct their importance signals. AdaLoRA~\citep{zhang2023adalora} scores singular-value triplets by parameter--gradient sensitivity, whereas IGU-LoRA~\citep{cui2026igulora} uses integrated gradients with uncertainty-aware smoothing. La-LoRA~\citep{gu2026lalora} relies on norm growth and layer contribution; SoRA~\citep{ding2023sora} and L1RA~\citep{singh2025l1ra} employ sparse gates, with $L_1$ regularization in L1RA; and ALoRA~\citep{liu2024alora} estimates importance through validation-time per-rank ablation.
These methods support non-uniform allocation, but their importance signals do not directly tie rank reallocation to a local loss-perturbation estimate derived from observed training movement.

LIRA instead directly adjusts layer-wise effective rank in standard $(A,B)$ LoRA and scores the factor displacements $\Delta A$ and $\Delta B$ observed between two reallocations.
Its local quadratic surrogate combines a first-order gradient--displacement interaction with a curvature term, takes the resulting perturbation magnitude, and normalizes it by the effective rank to obtain a cross-layer per-rank score.
Thus, each rank update can be traced to an interpretable estimate of local loss-perturbation magnitude rather than only to a proxy signal.
\section{Preliminaries}
\label{sec:prelim}

\paragraph{LoRA module.}
Consider a frozen pretrained linear map with weights $W^{(0)}\in\mathbb{R}^{d_{\mathrm{out}}\times d_{\mathrm{in}}}$.
Low-rank adaptation~\citep{hu2022lora} adds a trainable increment $\Delta W = B A$ with
$A\in\mathbb{R}^{r_{\max}\times d_{\mathrm{in}}}$ and $B\in\mathbb{R}^{d_{\mathrm{out}}\times r_{\max}}$, where $r_{\max}$ is a per-layer \emph{maximum} rank cap.
The forward pass uses
\begin{equation}
  h = W^{(0)}x + B A x.
\end{equation}
Unless stated otherwise, we apply one LoRA module per adapted linear projection. There are $L$ such modules, indexed by $i\in\{1,\ldots,L\}$, each with its own $(A_i,B_i)$ and analogous notation indexed by $i$.

\paragraph{Effective rank and masking.}
We keep full trainable factors $(A_i,B_i)$ of dimension $r_{\max}$, but only the leading $r_i$ components contribute to the forward pass at time $t$.
Concretely, let $A_{i}^{(t)}[1{:}r_i^{(t)},:]$ denote the first $r_i^{(t)}$ rows of $A_i$ and $B_{i}^{(t)}[:,1{:}r_i^{(t)}]$ the first $r_i^{(t)}$ columns of $B_i$; here and below, indices start at $1$.
All remaining rows/columns are masked out (not used in $B_iA_i$) but may become active later when $r_i^{(t)}$ increases.
We require
\begin{equation}
  r_{\min} \le r_i^{(t)} \le r_{\max}, \qquad
  R^{(t)} := \sum_{i=1}^{L} r_i^{(t)}.
\end{equation}
Here, $R^{(t)}$ is the realized total effective rank, whereas $\mathcal{B}^{(t)}$ is the effective-rank budget upper bound; thus, $R^{(t)}\le\mathcal{B}^{(t)}$.
For a controlled comparison to standard LoRA that uses a uniform rank $r_{\mathrm{base}}$ on the same modules, one sets the initial effective ranks and any fixed budget so that $R^{(0)}=\mathcal{B}^{(0)}=L\,r_{\mathrm{base}}$ (or matches active LoRA parameters under an equivalent counting convention).

\paragraph{Training objective.}
Let $\mathcal{L}$ be the downstream loss (e.g., cross-entropy).
Only LoRA parameters and task heads (if any) are updated; $W^{(0)}$ stays frozen.
We use the same optimizer as standard LoRA training; the following section describes periodic \textbf{DELTA} reallocation on top of ordinary gradient steps.
\section{Dynamic Loss-Informed Rank Allocation (LIRA)}
\label{sec:method}

As motivated in the introduction, dynamic rank allocation should do more than move budget from one layer to another: each reallocation should be tied to an explicit local estimate of how recent parameter movement perturbs the loss.
We therefore base the allocation signal on LoRA-factor displacements observed between successive reallocations. The allocator assigns more effective capacity in the next round to layers with a larger estimated local loss-perturbation magnitude.
We instantiate LIRA on top of standard $(A,B)$ LoRA with an effective-rank mask.
Let $K_{\mathrm{step}}$ denote the actual reallocation interval used by the training loop, measured in optimizer steps.
\textbf{DELTA} is applied every $K_{\mathrm{step}}$ optimizer steps:
\textbf{D}isplacement-\textbf{E}valuated \textbf{L}oss-\textbf{T}rajectory \textbf{A}llocation.

Each \textbf{DELTA} pass starts from interval-relative displacements $\Delta A$ and $\Delta B$, estimates layer-wise local loss-perturbation magnitudes with a quadratic surrogate, normalizes them to per-rank scores $S_i^{(t)}$, and obtains integer effective ranks $\{r_i^{(t)}\}$ under the effective-rank budget $\mathcal{B}^{(t)}$.
Scoring and allocation therefore revolve around the same loss-perturbation score. Capacity is controlled through \emph{layer-wise effective ranks}---the number of rank components active in the forward pass---which keeps the method compatible with existing training stacks and matched-budget LoRA baselines.

\subsection{Overview and Algorithm}
\label{sec:overview}

\begin{figure*}[t]
  \centering
  \includegraphics[width=0.98\textwidth]{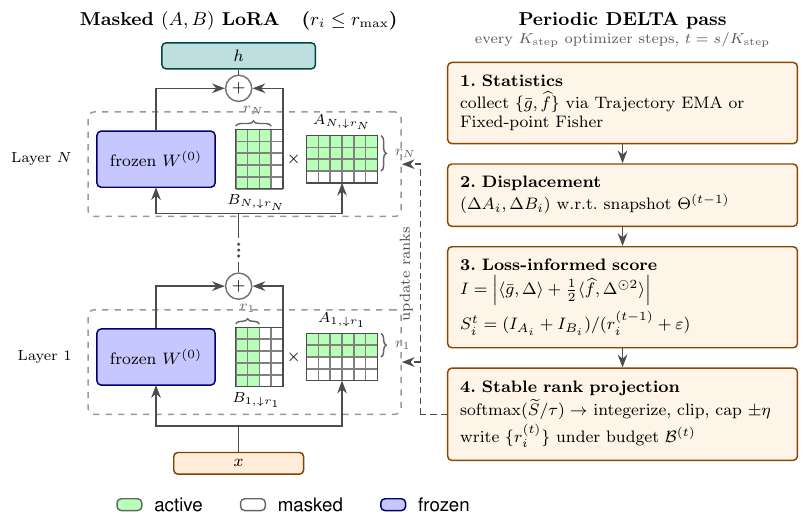}
\caption{LIRA overview.
  Standard masked LoRA updates (left) are interleaved every $K_{\mathrm{step}}$ optimizer steps with a DELTA pass (right), which converts interval displacements into local loss-perturbation magnitudes that combine first- and second-order terms, rank-normalized scores $S_i^{(t)}$, and discrete ranks under $\mathcal{B}^{(t)}$.}
  \label{fig:lira-overview}
\end{figure*}

Training proceeds for $n$ optimizer steps $s=1,\ldots,n$.
After each step we apply a standard forward pass, backward pass, and optimizer update on all LoRA parameters under the current mask.
Every $K_{\mathrm{step}}$ optimizer steps, immediately after an update at $s\in\{K_{\mathrm{step}},2K_{\mathrm{step}},\ldots\}$, we run a \textbf{DELTA} pass indexed by $t=s/K_{\mathrm{step}}$. The pass collects local gradient statistics, forms interval displacements $(\Delta A,\Delta B)$ on the preceding active-rank support, computes $\{S_i^{(t)}\}$, and projects integer ranks $\{r_i^{(t)}\}$ under budget $\mathcal{B}^{(t)}$ (Algorithm~\ref{alg:lira}).
Between DELTA passes, masked coordinates receive zero effective gradient; when $r_i$ increases, newly activated rows/columns may be zero-initialized or kept from the original initialization.
Figure~\ref{fig:lira-overview} summarizes the masked $(A,B)$ stack and one DELTA pass.

\begin{algorithm}[t]
\caption{LIRA fine-tuning with periodic DELTA}
\label{alg:lira}
\begin{algorithmic}[1]
  \REQUIRE $r_{\max}$, $r_{\min}$, budget $\mathcal{B}$ (fixed or scheduled as $\mathcal{B}^{(t)}$); reallocation interval $K_{\mathrm{step}}$ (optimizer steps); $L$ LoRA modules; initial ranks $\{r_i^{(0)}\}$; temperature $\tau$; step cap $\eta$;
$\varepsilon>0$
\STATE Initialize $(A_i,B_i)$; set effective ranks $\{r_i^{(0)}\}$; snapshot active factors $\Theta^{(0)}\leftarrow\{(A_i,B_i)_{\downarrow r_i^{(0)}}\}$
\FOR{$s=1,\ldots,n$}
  \STATE Forward/backward/optimizer update under current masks
  \IF{$s \bmod K_{\mathrm{step}} = 0$}
    \STATE $t\leftarrow s/K_{\mathrm{step}}$; set $\mathcal{B}^{(t)}$ (else $\mathcal{B}^{(t)}\leftarrow\mathcal{B}$)
  \STATE Collect $\{\bar g,\hat f\}$ via Trajectory EMA \textbf{or} Fixed-point Fisher (score estimation)
    \FOR{each layer $i=1,\ldots,L$}
      \STATE $\Delta A_i,\Delta B_i \leftarrow$ current factors minus $\Theta^{(t-1)}$, both on support $r_i^{(t-1)}$
      \STATE $I_{A_i},I_{B_i}\leftarrow$ local loss-perturbation magnitudes under $(\bar g,\hat f)$
      \STATE $S_i^{(t)}\leftarrow (I_{A_i}+I_{B_i})/(r_i^{(t-1)}+\varepsilon)$
    \ENDFOR
    \STATE Optionally $z$-score $\{S_i^{(t)}\}$ to obtain $\widetilde S^{(t)}$
    \STATE $w^{(t)}\leftarrow\mathrm{softmax}(\widetilde S^{(t)}/\tau)$; $\widehat r_i^{(t)}\leftarrow w_i^{(t)}\,\mathcal{B}^{(t)}$
    \STATE Integerize $\{\widehat r_i^{(t)}\}$ (largest remainders); clip to $[r_{\min},r_{\max}]$
    \STATE Cap per-layer change by $\pm\eta$; cancel decreases if $R^{(t)}$ would fall below $R_{\min}^{(t)}$
    \STATE Apply decreases first, then fund increases from freed/\allowbreak residual budget (rank projection)
    \STATE Write $\{r_i^{(t)}\}$; refresh snapshot $\Theta^{(t)}$; optionally anneal $\tau\leftarrow\max(\tau_{\min},\rho\tau)$
  \ENDIF
\ENDFOR
\end{algorithmic}
\end{algorithm}

\subsection{Interval Displacement and Loss-Perturbation Scoring}
\label{sec:scores}

\paragraph{Scoring objective.}
Under a fixed effective-rank budget, the score for layer $i$ answers a concrete question: \emph{what is the estimated magnitude of the local loss perturbation associated with this layer's recent movement on its effective ranks?}
DELTA defines $S_i^{(t)}$ from a loss-perturbation perspective.
Under a local second-order Taylor approximation, $S_i^{(t)}$ estimates the unsigned magnitude associated with layer $i$'s recent parameter movement. The first-order term captures the gradient--displacement interaction, and the second-order term captures the joint effect of movement magnitude and local loss curvature. Thus, \emph{more interpretable} is meant in a specific, relative sense: the score is expressed in loss units and computed from observed factor movement rather than an indirect proxy, so at every DELTA pass the basis for adjusting $r_i$ can be decomposed into these two terms and audited against the layer's recent movement.

\paragraph{Interval-relative increments.}
Immediately before the $t$th DELTA pass, the effective rank is $r_i^{(t-1)}$, and the snapshot from the preceding pass is restricted to the same active support.
Let $t^-$ denote the parameter state immediately before the pass, and let $\Theta^{(t-1)}$ be the active-factor snapshot saved at the preceding pass.
The increment between successive DELTA passes is
\begin{equation}
  \begin{aligned}
  \Delta A_i^{(t)} &= A_{i,\downarrow r_i^{(t-1)}}^{(t^-)}
  - \Theta_{A,i}^{(t-1)}, \\
  \Delta B_i^{(t)} &= B_{i,\downarrow r_i^{(t-1)}}^{(t^-)}
  - \Theta_{B,i}^{(t-1)}.
  \end{aligned}
  \label{eq:delta}
\end{equation}
$\Delta A_i^{(t)}$ and $\Delta B_i^{(t)}$ record the actual movement over the preceding $K_{\mathrm{step}}$ optimizer steps on the ranks active before pass $t$.

\paragraph{From loss perturbation to a readable score.}
To estimate the effect of this finite displacement on $\mathcal{L}$, we use a local quadratic model following Taylor-based sensitivity analysis in compression and pruning~\citep{lecun1990obd,molchanov2019importance,kim2024squeezellm}:
\begin{equation}
  \mathcal{L}(\theta+\delta) \approx \mathcal{L}(\theta) + g^{\top}\delta + \tfrac{1}{2}\,\delta^{\top}\!\operatorname{diag}(\widehat{f})\,\delta,
  \label{eq:taylor}
\end{equation}
where $g$ is the local gradient and $\widehat{f}$ is a diagonal Fisher-style curvature proxy (an elementwise second-moment statistic).
Applied blockwise to the active part of $A_i$, with Frobenius inner product $\langle U,V\rangle=\sum_{p,q}U_{pq}V_{pq}$, we define the unsigned local loss-perturbation magnitude associated with layer $i$'s recent $A$-movement as
\begin{equation}
  I_{A_i}^{(t)} \approx \left|
  \big\langle \overline{g}_{A_i}^{\,(t)},\, \Delta A_i^{(t)} \big\rangle
  + \frac{1}{2} \big\langle \widehat{f}_{A_i}^{\,(t)},\, (\Delta A_i^{(t)})^{\odot 2} \big\rangle
  \right|,
  \label{eq:IA}
\end{equation}
and analogously $I_{B_i}^{(t)}$ on the active block of $B_i$.
The absolute value makes $I_{A_i}^{(t)}$ and $I_{B_i}^{(t)}$ unsigned magnitudes rather than signed loss increases or decreases.
We score magnitude because either sign indicates that the loss is locally sensitive to movement on the layer's active components: a large predicted decrease marks adaptation in progress, and a large predicted increase marks movement through a high-curvature region, both of which warrant capacity; a near-zero perturbation instead indicates that recent movement barely engages the loss, so the layer's rank can be reallocated.
This use of unsigned magnitudes mirrors Taylor-based importance criteria in pruning~\citep{lecun1990obd,molchanov2019importance}.
Combining them and dividing by the effective rank during the preceding interval yields a \emph{per-rank} score comparable across layers:
\begin{equation}
  S_i^{(t)} = \frac{I_{A_i}^{(t)} + I_{B_i}^{(t)}}{r_i^{(t-1)} + \varepsilon}.
  \label{eq:Si}
\end{equation}
Higher $S_i^{(t)}$ means that layer $i$ has a larger local loss-perturbation magnitude \emph{per effective rank} during the preceding interval; DELTA therefore assigns it a larger rank target in the current round.
\paragraph{Estimating gradients and curvature.}
To estimate $\overline{g}$ and $\widehat{f}$, we treat gradients from successive mini-batches along the fine-tuning trajectory as approximate samples from the same underlying data distribution.
Under this assumption, mean gradients and elementwise squared gradients can be accumulated from gradients produced naturally during training; when an estimate closer to a fixed empirical Fisher at a reallocation point is desired, additional sampling near that point is also possible.
We use two implementations:
\textbf{(A) Trajectory EMA.}
Between DELTA passes, maintain running means and elementwise squared gradients on LoRA parameters (e.g., reuse Adam/AdamW moments with implementation-specific scaling).
Overhead is small and the statistics evolve along the optimization path.
\textbf{(B) Fixed-point Fisher window.}
At a selected DELTA step, freeze LoRA parameters, run $N_{\mathrm{stat}}$ extra forward/backward passes without updates, average gradients and squared gradients at that point, then resume training.
This is closer to a textbook empirical Fisher estimate but requires additional forward/backward computation.
A short warm-up before the first DELTA pass helps stabilize the moment estimates.

\subsection{Rank Projection under an Allocation Budget}
\label{sec:allocation}

Given $\{S_i^{(t)}\}$, DELTA converts these scores into integer ranks under the effective-rank budget upper bound $\mathcal{B}^{(t)}$.
Since score standardization and the softmax preserve the ordering of the scores, layers with higher local loss-perturbation scores receive larger rank targets.

\paragraph{Softmax targets under a budget.}
We may first standardize $\{S_i^{(t)}\}$ across layers ($z$-score), denoting the standardized score by $\widetilde S_i^{(t)}$; when standardization is disabled, let $\widetilde S_i^{(t)}=S_i^{(t)}$.
We then define
\begin{equation}
  w_i^{(t)} = \frac{\exp(\widetilde S_i^{(t)}/\tau)}{\sum_{j=1}^{L}\exp(\widetilde S_j^{(t)}/\tau)},
\end{equation}
and continuous targets $\widehat{r}_i^{\,(t)} = w_i^{(t)}\, \mathcal{B}^{(t)}$.
The temperature $\tau$ trades off between uniform allocation (large $\tau$) and concentrating budget on high-scoring layers (small $\tau$); we optionally anneal $\tau$ after each DELTA pass.

\paragraph{Discrete projection and bounded rank moves.}
We first integerize $\{\widehat{r}_i^{\,(t)}\}$ by flooring and assigning residual units to the largest fractional parts, then clip to $[r_{\min},r_{\max}]$.
Direct jumps to these targets can still be abrupt, so each layer's change is capped by a step size $\eta$  (which may be set directly or, in one implementation, is set from $r_{\min}$, the initial rank, and the expected number of DELTA passes).
Decreases are applied first; residual effective-rank budget under $\mathcal{B}^{(t)}$ then funds increases in a round-robin over layers whose continuous targets exceed their current ranks.
To prevent the realized total rank from falling too low, we impose
\begin{equation}
  R_{\min}^{(t)}=\max(\gamma\mathcal{B}^{(t)},Lr_{\min}),
\end{equation}
and cancel the weakest tentative decreases whenever the projected $R^{(t)}$ would violate this floor.

\paragraph{Global budget scheduling.}
The effective-rank budget upper bound $\mathcal{B}^{(t)}$ may also vary over training (e.g., cosine decay from an initial budget to a target value), playing a role similar to the descending budget scheduler in AdaLoRA~\citep{zhang2023adalora}. Within each budget level, layer-wise splits are still determined by $S_i^{(t)}$.
If no scheduler is used, $\mathcal{B}^{(t)}$ remains the fixed constant $\mathcal{B}$ throughout fine-tuning.

\begin{table*}[t]
\centering
\small
\setlength{\tabcolsep}{2pt}
\begin{tabular*}{\textwidth}{@{\extracolsep{\fill}}lcccccccc@{}}
\toprule
Method & Active Params. & SST-2 & MRPC & CoLA & QNLI & RTE & STS-B & Avg. \\
\midrule
Full FT & $125$M & $\underline{94.8}$ & $\mathbf{90.2}$ & $63.6$ & $92.8$ & $78.7$ & $\mathbf{91.2}$ & $85.2$ \\
LoRA~\citep{hu2022lora} & $0.3$M & $94.0$ & $88.7$ & $63.4$ & $92.4$ & $78.6$ & $90.5$ & $84.6$ \\
AdaLoRA~\citep{zhang2023adalora} & $0.3$M & $\underline{94.8}$ & $88.8$ & $63.2$ & $\underline{92.9}$ & $79.0$ & $\underline{91.1}$ & $85.0$ \\
DoRA~\citep{liu2024dora} & $0.3$M & $93.5$ & $87.6$ & $64.2$ & $91.0$ & $79.7$ & $91.0$ & $84.5$ \\
La-LoRA~\citep{gu2026lalora} & $0.3$M & $\underline{94.8}$ & $88.0$ & $\underline{65.0}$ & $\underline{92.9}$ & $\underline{80.4}$ & $91.0$ & $\underline{85.4}$ \\
\midrule
LIRA (ours) & $0.3$M$^{\dagger}$ & $\mathbf{95.3}$ & $\underline{90.1}$ & $\mathbf{65.3}$ & $\mathbf{93.0}$ & $\mathbf{81.5}$ & $\underline{91.1}$ & $\mathbf{86.0}$ \\
\bottomrule
\end{tabular*}
\caption{RoBERTa-base GLUE development results. LIRA entries report five-seed means, and all values are shown to one decimal place; baseline values are taken from Table~5 of La-LoRA~\citep{gu2026lalora}. Best results are boldfaced, and second-best results are underlined. $^{\dagger}$ The reported count is the active forward-pass parameter budget, matched to uniform LoRA with effective rank $r=8$; LIRA additionally stores masked factors up to $r_{\max}$ per layer. The Avg. column is the arithmetic mean of the six task means. The complete main-results table is provided in the supplementary material.}
\label{tab:glue-main}
\end{table*}
\section{Experiments}
\label{sec:experiments}

We evaluate LIRA on RoBERTa-base GLUE following the protocol reported for La-LoRA, a recent adaptive-rank LoRA variant.
Unless otherwise stated, the main results use the \textbf{Trajectory EMA} estimator for local loss-perturbation scores; we also evaluate the \textbf{Fixed-point Fisher window} estimator as an alternative.

\subsection{Experimental Setup}
\label{sec:exp-setup}

\paragraph{Protocol and baselines.}
We follow the RoBERTa-base~\citep{liu2019roberta} GLUE~\citep{wang2018glue} setting reported by La-LoRA~\citep{gu2026lalora}, which adopts the LoRA configuration in Table~9 of \citet{hu2022lora}. We use AdamW with linear decay and a warmup ratio of $0.06$, a maximum sequence length of $512$, and adapters on the query and value projections; MRPC, RTE, and STS-B do not use MNLI-checkpoint initialization.
As in La-LoRA, we run five random seeds and report the mean over seeds; within each run we take the best development epoch.
We report Matthews correlation on CoLA, Pearson correlation on STS-B, and accuracy on the remaining tasks.
Results for Full Fine-Tuning, LoRA, AdaLoRA, DoRA, and La-LoRA are taken from Table~5 of \citet{gu2026lalora}.

\paragraph{LIRA configuration.}
We match the cited uniform LoRA setting with base rank $r_{\mathrm{base}}=8$: DELTA reallocates across the $L$ adapted modules under the effective-rank budget upper bound $\mathcal{B}=L\,r_{\mathrm{base}}$. After discrete projection, the realized total effective rank $R$ may be lower than this upper bound.
Each layer stores factors up to a capacity $r_{\max}\ge r_{\mathrm{base}}$ (typically $r_{\max}=16$), while only the leading effective ranks enter the forward pass.
DELTA uses a task-specific optimizer-step interval $K_{\mathrm{step}}$.
Task intervals are specified by an epoch-length multiplier $c_{\mathrm{epoch}}$.
Let each data-parallel replica contain $E$ optimizer steps per epoch and let $W$ denote the data-parallel world size; then
\begin{equation}
K_{\mathrm{step}}=\max\{1,\lfloor c_{\mathrm{epoch}} E W\rfloor\}.
\end{equation}
Unless noted, local loss-perturbation scores use Trajectory EMA with the second-order diagonal term enabled.

\subsection{Main Results on GLUE}
\label{sec:main-results}

Table~\ref{tab:glue-main} reports the main comparison.
With a matched effective-rank budget, LIRA improves the six-task average by approximately $1.4$ points over the reported LoRA result and $0.6$ points over the reported La-LoRA result. Relative to LoRA, the largest gains are on RTE and CoLA; relative to La-LoRA, they are on MRPC and RTE.

\begin{figure*}[t]
\centering
\includegraphics[width=\textwidth]{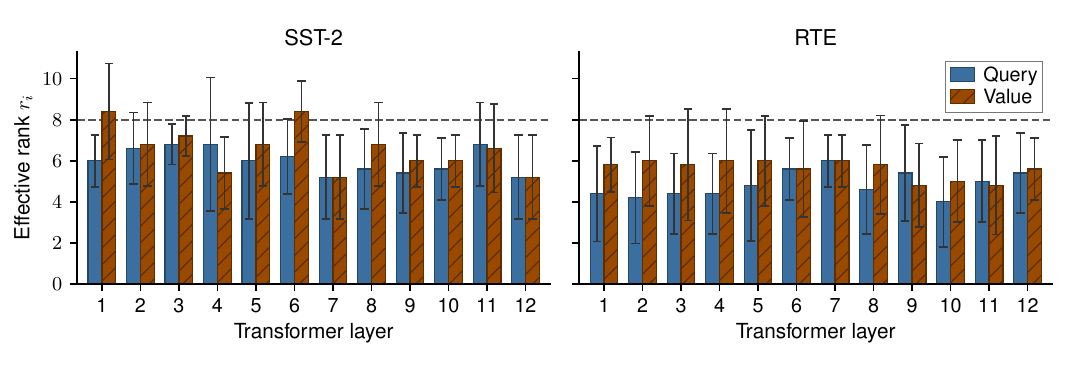}
\caption{Final effective-rank profiles after DELTA (Trajectory EMA; five-seed means) on SST-2 and RTE.
Each bar represents the query or value adapter at one RoBERTa-base transformer layer; the dashed line marks the uniform baseline $r_{\mathrm{base}}=8$.
Both tasks reallocate away from a uniform profile under the shared budget upper bound $L\,r_{\mathrm{base}}$, while the realized total effective rank may remain below this bound after discrete projection.}
\label{fig:final-rank-dist}
\end{figure*}

Figure~\ref{fig:final-rank-dist} shows the corresponding end-of-training allocations on SST-2 and RTE.
Query and value adapters receive different ranks within the same block, and both profiles are non-uniform relative to $r_{\mathrm{base}}=8$.
The task-specific allocation patterns also differ: SST-2 concentrates relatively more rank in middle and early value adapters, whereas RTE assigns lower ranks overall and redistributes them differently across depth. This supports the premise that adapter importance depends on both layer and task, rather than following a single shared ranking over blocks.

\subsection{Trajectory EMA vs.\ Fixed-Point Fisher Window}
\label{sec:ema-vs-fisher}

The Method section introduces two estimators of $(\overline{g},\widehat{f})$.
Table~\ref{tab:ema-fisher} compares them under the same budgets and reallocation schedules.

\begin{table}[t]
\centering
\small
\begin{tabular*}{\columnwidth}{@{\extracolsep{\fill}}lcc@{}}
\toprule
Task & Trajectory EMA & Fixed-point Fisher \\
\midrule
SST-2 & $95.25$ & $\mathbf{95.32}$ \\
MRPC & $90.05$ & $\mathbf{90.29}$ \\
CoLA & $65.26$ & $\mathbf{65.86}$ \\
QNLI & $92.96$ & $\mathbf{93.01}$ \\
RTE & $\mathbf{81.52}$ & $80.94$ \\
STS-B & $\mathbf{91.11}$ & $91.07$ \\
Average & $86.03$ & $\mathbf{86.08}$ \\
\bottomrule
\end{tabular*}
\caption{Trajectory EMA and Fixed-point Fisher under the same local loss-perturbation score, effective-rank budget, and five-seed GLUE protocol. Values are five-seed means reported to two decimal places; best results are boldfaced, and task metrics follow Table~\ref{tab:glue-main}.}
\label{tab:ema-fisher}
\end{table}

\begin{table*}[t]
\centering
\small
\begin{tabular*}{\textwidth}{@{\extracolsep{\fill}}lccccccc@{}}
\toprule
Variant & SST-2 & MRPC & CoLA & QNLI & RTE & STS-B & Avg. \\
\midrule
w/o 2nd-order term & $94.89$ & $89.56$ & $64.50$ & $92.63$ & $80.14$ & $90.80$ & $85.42$ \\
w/o per-effective-rank averaging & $94.91$ & $89.61$ & $64.21$ & $92.71$ & $78.70$ & $90.91$ & $85.17$ \\
LIRA (full) & $\mathbf{95.25}$ & $\mathbf{90.05}$ & $\mathbf{65.26}$ & $\mathbf{92.96}$ & $\mathbf{81.52}$ & $\mathbf{91.11}$ & $\mathbf{86.03}$ \\
\bottomrule
\end{tabular*}
\caption{Ablation of LIRA's key components. ``w/o'' removes the stated design; LIRA (full) retains all components. Values are means over five random seeds; best results are boldfaced; task metrics match Table~\ref{tab:glue-main}.}
\label{tab:key-component-ablations}
\end{table*}

The two estimators give similar averages ($86.03$ versus $86.08$), with Fixed-point Fisher slightly higher.
Although Fisher has a slightly higher average, Table~\ref{tab:glue-main} uses Trajectory EMA as the representative estimator. It reuses statistics already available along the optimization path and avoids extra frozen forward/backward passes at each DELTA step.
Fixed-point Fisher remains useful when a closer empirical-Fisher estimate at the reallocation point is preferred and the additional computation is acceptable.

\subsection{Ablation of Key Components}
\label{sec:key-component-ablations}

LIRA computes layer-wise local loss-perturbation scores from LoRA-factor displacements within each DELTA reallocation interval.
We examine two key designs: the diagonal Fisher-style curvature term that supplies local second-order information in Trajectory EMA, and per-effective-rank averaging, $1/(r_i^{(t-1)}+\varepsilon)$, which converts total layerwise magnitude into an average per effective rank.
For each ablation, we remove only one design while holding the model, optimizer, effective-rank budget, and DELTA schedule fixed relative to full LIRA.

\paragraph{Second-order curvature.}
Table~\ref{tab:key-component-ablations} shows that full LIRA has higher five-seed means on all six task metrics than the variant without the second-order curvature term. The average gain is $0.61$ pp, with gains of $1.38$ pp on RTE and $0.76$ pp on CoLA.
The two terms encode complementary local information: the first-order term measures gradient--displacement interaction, whereas the second-order term incorporates the curvature of the coordinates traversed by the displacement.
Without the latter, displacements with similar first-order magnitude but different local curvature receive similar scores.

\paragraph{Per-effective-rank averaging.}
Full LIRA also has higher five-seed means on all six task metrics than the variant without per-effective-rank averaging, with an average gain of $0.86$ pp and gains of $2.82$ pp and $1.05$ pp on RTE and CoLA, respectively.
Layerwise magnitude is summed over active factors, so a layer with more active factors can receive a larger total score purely because it already has more rank. This can create a self-reinforcing allocation pattern.
Dividing by the effective rank converts total magnitude into an average contribution per effective rank, reducing this mechanical dependence on existing capacity.

\section{Conclusion}
\label{sec:conclusion}

We presented LIRA, a dynamic-rank LoRA method designed to make online rank reallocation \emph{more interpretable}.
To this end, LIRA combines a first-order gradient--displacement interaction with a second-order curvature term, then evaluates each layer through the unsigned magnitude of a local loss perturbation over its recent LoRA-factor movements.
This loss-perturbation view makes the basis for each rank change interpretable term by term, while periodic DELTA passes reallocate effective ranks under matched effective-rank budgets and remain compatible with standard $(A,B)$ adapters.
On RoBERTa-base GLUE under the La-LoRA/LoRA protocol, LIRA improves over the uniform-LoRA and adaptive-rank baselines reported under this protocol. Trajectory EMA matches Fixed-point Fisher ($86.0$ vs.\ $86.1$) without the latter's extra statistic-window computation.
Our interpretability claim is scoped to the construction of the allocation signal: each rank change is traceable to a term-wise local loss-perturbation estimate.
Overall, LIRA shows how adaptive PEFT can couple effective-rank allocation with local loss-perturbation signals derived from observed training movement.
\bibliography{references}
\end{document}